\documentclass{pbmlarxiv} \pdfoutput=1

\usepackage{color}
\usepackage{colortbl}
\usepackage{multirow}
\usepackage{tabularx}
\usepackage[table,xcdraw]{xcolor}
\usepackage{pstricks}
\usepackage{booktabs}

\begin{document}


\title{Text Summarization of Czech News Articles Using Named Entities}


\institute{}{Faculty of Electrical Engineering, CTU in Prague, Prague, Czech Republic}
\author{firstname=Petr, surname=Marek,
  corresponding=yes,
  email={marekp17@fel.cvut.cz},
  address={Faculty of Electrical Engineering\\
           Czech Technical University in Prague\\
           Technick\'a 2\\
           166 27 Praha 6 - Dejvice, Czech Republic}}
\author{firstname=\v{S}t\v{e}p\'an, surname=M\"{u}ller}
\author{firstname=Jakub, surname=Konr\'ad}
\author{firstname=Petr, surname=Lorenc}
\author{firstname=Jan, surname=Pichl}
\author{firstname=Jan, surname=\v{S}ediv\'y}

\shorttitle{Text Summarization of Czech News Articles Using NE}
\shortauthor{P. Marek et al.}

\PBMLmaketitle


\begin{abstract}
The foundation for the research of summarization in the Czech language was laid by the work of \citet{straka2018sumeczech}. They published the SumeCzech, a  large Czech news-based summarization dataset, and proposed several baseline approaches. However, it is clear from the achieved results that there is a large space for improvement.

In our work, we focus on the impact of named entities on the summarization of Czech news articles. First, we annotate SumeCzech with named entities. We propose a new metric \linebreak ROUGE\textsubscript{NE} that measures the overlap of named entities between the true and generated summaries, and we show that it is still challenging for summarization systems to reach a high score in it.

We propose an extractive summarization approach \emph{Named Entity Density} that selects a sentence with the highest ratio between a number of entities and the length of the sentence as the summary of the article. The experiments show that the proposed approach reached results close to the solid baseline in the domain of news articles selecting the first sentence. Moreover, we demonstrate that the selected sentence reflects the style of reports concisely identifying \emph{to whom}, \emph{when}, \emph{where}, and \emph{what} happened. We propose that such a summary is beneficial in combination with the first sentence of an article in voice applications presenting news articles.

We propose two abstractive summarization approaches based on Seq2Seq architecture. The first approach uses the tokens of the article. The second approach has access to the named entity annotations. The experiments show that both approaches exceed state-of-the-art results previously reported by \citet{straka2018sumeczech}, with the latter achieving slightly better results on SumeCzech's out-of-domain testing set.
\end{abstract}

\section{Introduction}

Automatic text summarization is an important task of natural language understanding. The goal is to describe a text accurately, be it a news article, a web page, or a paragraph of a book, using shorter text. The shorter text can be in the form of a paragraph, sentence, or even a few words. Automatic text summarization is a challenging problem for automatic systems because they have to excel in multiple areas at once. They have to understand the meaning of the original text, understand which passages are important and which can be excluded, and generate meaningful and grammatically correct summarizations.

In this work, we focus on the summarization of Czech news articles by a \linebreak one-sentence summary. We can also describe this task as the automatic creation of a headline for a given text. We use the SumeCzech dataset \citep{straka2018sumeczech} for our experiments. 

Additionally, we explore the influence of the named entities on text summarization. We use SpaCy’s named entity recognition (NER) model, trained on a \linebreak CoNLL-based extended CNEC 2.0 dataset \citep{11858/00-097C-0000-0023-1B22-8}, to label SumeCzech with named entities to create additional features. We publish the annotations to promote the replication of results and to enable further research.

We use the annotations as a foundation for our newly proposed \linebreak \emph{Named Entity Density}. The method selects the sentence with the highest ratio of the number of entities to the sentence length as the summary of the article. We show that our proposed method achieves nearly as good results in the automatic evaluation as the hard-to-beat baseline in the news domain that selects the first sentence. \citet{nenkova2005automatic} shows that the baseline selecting the first sentence is a strong baseline because authors tend to summarize the main points of an article in the first sentence, especially in the news domain. We also show that sentences selected by \linebreak \emph{Named Entity Density} possess a high information value mentioning \emph{to whom}, \emph{where}, \emph{when}, and \emph{what} happened. This structure resembles the style of reports that concisely identifies and examines issues, events, or findings that have happened. We propose that such a summary is useful in voice applications presenting news. Voice application can present the summary formed out of the first sentence of a news article first and continue with the sentence selected by \emph{Named Entity Density} if a user requests additional information.

We also propose two abstractive methods that can construct a novel sentence as a summary. They are based on the Seq2Seq architecture, initially used for machine translation. The first method uses the text of the article only. The second method uses additional annotations created by the name entity recognition system as input features. Our experiments show that both models achieved state-of-the-art results in SumeCzech's task to summarize the headline from the text of the article. We also show that the named entities added as an additional input feature improve the ability of the model to generalize to the out-of-domain data.

Finally, we propose a new metric Rogue\textsubscript{NE}, which measures the overlap of named entities in the target and generated summaries. Poor results of the experiments in Rogue\textsubscript{NE} show that summarization of entities still poses many challenges, and this task has not been solved yet.

\section{Related Work}
\citet{allahyari2017text} provide a brief survey of text summarization. In general, we divide text summarization algorithms into two categories, \emph{extractive} and \emph{abstractive}. 

\subsection{Extractive Summarization}

Extractive summarization algorithms choose pieces from the original text, usually sentences, and combine them to form a summary. From a high-level perspective, most extractive summarizers follow the same two steps: First, score all sentences. Then, pick \(N\) sentences with the highest score. The main difference between individual extractive methods is how they score sentences. The advantage of extractive methods is that no matter how simple the method is, it always produces syntactically correct sentences, even though they may not be useful summaries. On the other hand, there is a disadvantage too. Extractive summarizers are limited in what they can predict by the sentences of the source text. Thus, more elaborate summaries are out of their reach.

\citet{mihalcea2004textrank} introduce a Textrank algorithm, a graph-based ranking model. It creates a graph of sentences based on their overlap. It chooses the most important sentences according to the created graph. \citet{pal2014approach} propose a summarization algorithm that derives the relevance of the sentences within the text using the Simplified Lesk algorithm and the WordNet online database. \citet{kaageback2014extractive} propose using continuous vector representations for semantically aware representations of sentences for summarization. \citet{zhang2016extractive} develop convolutional neural networks that learn sentence features and perform sentence ranking. The latest results are achieved by \citet{liu2019fine}. They apply the BERT model \citep{devlin2018bert} to extractive summarization.

Especially relevant works for our research are those working with named entities. \citet{nobata2002summarization} introduce named entity tagging and pattern discovery to a summarization system based on a sentence extraction technique. \citet{hassel2003exploitation} integrates a Named Entity tagger into the SweSum summarizer for Swedish newspaper texts. \citet{filatova2004event} propose a summarization technique using a set of features based on low-level, atomic events that describe the relationships between important actors in a document or in a set of documents. The extraction of atomic events relies on a noun phrase and named entity recognition \citep{hatzivassiloglou2003domain}. \citet{jabeen2013named} apply named entity recognition for summarization of tweets. \citet{schulze2016entity} present EntityRank, a multidocument graph-based summarization algorithm that is solely based on named entities. They apply it to texts from the medical domain successfully. \citet{khademi2020persian} propose an unsupervised method for summarizing Persian texts that use a named entity recognition system. Their method consists of three phases: training a supervised NER model, recognizing named entities in the text, and generating a summary.

\subsection{Abstractive Summarization}

Abstractive summarizers generate summarizations consisting of novel sentences that were not part of the original text. Abstractive summarization algorithms are usually more complex because they have to understand the input text, find the most relevant passages, and generate syntactically correct sentences as summarization. Such a task is nearly impossible for hand-written rules. However, the recent advance of machine learning and, in particular, neural networks makes abstractive summarization possible. Moreover, neural networks represent the current state-of-the-art in abstractive summarization.

\citet{nallapati2016abstractive} models abstractive text summarization using Attentional Encoder-Decoder Recurrent Neural Networks. They propose several novel models that address critical problems in summarization that are not adequately modeled by the basic architecture, such as modeling keywords, capturing the hierarchy of sentence-to-word structure, and emitting rare or unseen words during the training time. \citet{liu2017generative} propose an adversarial process for abstractive text summarization. \citet{yao2018dual} propose a recurrent neural network-based Seq2Seq attentional model with a dual encoder including the primary and the secondary encoders. \citet{song2019abstractive} propose an LSTM-CNN based approach that can construct new sentences by exploring more fine-grained fragments than sentences, namely, semantic phrases. The proposed approach is composed of two main stages. The first stage extracts phrases from source sentences. The second stage generates text summaries using deep learning. \citet{liu2019text} apply BERT in text summarization and propose a general framework for both extractive and abstractive models. For abstractive summarization, they propose a new fine-tuning schedule that adopts different optimizers for the encoder and the decoder as a means of alleviating the mismatch between the two as the former is pretrained while the latter is not.

To the best of our knowledge, there is no work exploring the influence of named entities on the extractive summarization techniques, let alone in the Czech language.

\section{Dataset}
We use SumeCzech for experiments. SumeCzech is Czech news-based summarization dataset created by \citet{straka2018sumeczech}. It contains more than a million documents, consisting of a headline, several sentences long abstract, and a full text. The dataset was collected from various Czech news websites. We show the distribution of the websites in \autoref{tab:websites}.

SumeCzech is split into four parts. Three of them are the train, development, and test sets. Additionally, to simulate a real-life situation where a model is trained on data from one domain, and used on real data from other domains, \citet{straka2018sumeczech} created an out-of-domain (OOD) test set. OOD test set evaluates how models cope with news articles from domain never seen during training. They clustered the whole dataset into 25 clusters using K-Means on abstracts of the articles and selected one cluster as the OOD test set. The OOD test set contains approximately 4.5\% of all articles. The OOD testing set seems to contain news articles about concerts and festivals. The remaining articles were divided into train, development, and test sets in 86.5~:~4.5~:~4.5~ratio.

\begin{table}[]
\centering
\begin{tabular}{lrr}
\toprule
 \textbf{Website}        & \textbf{Number}    & \textbf{Percentage} \\ \midrule
 ceskenoviny.cz & 4,854     & 0.5\%      \\
 denik.cz       & 157,581   & 15.7\%     \\
 idnes.cz       & 463,192   & 46.2\%     \\
 lidovky.cz     & 136,899   & 13.7\%     \\
 novinky.cz     & 239,067   & 23.9\%     \\ \midrule
 Total          & 1,001,593 & 100.0\%    \\  
 \bottomrule
\end{tabular}%
\caption{The number of documents in SumeCzech from individual news websites}
\label{tab:websites}
\end{table}

\begin{table}[p!]
\centering
\begin{tabular}{rrrrr}
\toprule
\textbf{Entity Type} & \textbf{Train} & \textbf{Dev} & \textbf{Test} & \textbf{Test OOD} \\ \midrule
Numbers in addresses & 116,990        & 5,052        & 5,129         & 1,827             \\
Geographical names   & 5,271,938      & 282,440      & 285,307       & 212,637           \\
Institutions         & 4,488,357      & 222,524      & 234,147       & 250,555           \\
Media names          & 534,340        & 24,379       & 27,966        & 22,360            \\
Artifact names       & 2,367,532      & 118,938      & 108,811       & 196,009           \\
Personal names       & 7,991,790      & 406,938      & 395,867       & 646,556           \\
Time expressions     & 1,684,152      & 87,096       & 86,866        & 121,357           \\ \midrule
Total                & 22,455,099     & 1,147,367    & 1,144,093     & 1,451,301         \\ \bottomrule
\end{tabular}
\caption{Number of named entities in texts of SumeCzech's articles}
\label{tab:entities_text}
\end{table}

\begin{table}[p!]
\centering
\begin{tabular}{rrrrr}
\toprule
\textbf{Entity Type} & \textbf{Train} & \textbf{Dev} & \textbf{Test} & \textbf{Test OOD} \\ \midrule
Numbers in addresses & 331            & 18           & 12            & 26                \\
Geographical names   & 285,148        & 15,903       & 14,697        & 13,502            \\
Institutions         & 161,809        & 7,578        & 8,472         & 12,806            \\
Media names          & 9,088          & 371          & 420           & 718               \\
Artifact names       & 62,124         & 3,344        & 2,837         & 7,748             \\
Personal names       & 302,276        & 15,117       & 15,856        & 31,266            \\
Time expressions     & 14,400         & 760          & 838           & 1,127             \\ \midrule
Total                & 835,176        & 43,091       & 43,132        & 67,193         \\ \bottomrule
\end{tabular}
\caption{Number of named entities in headlines of SumeCzech's articles}
\label{tab:entities_headline}
\end{table}

\begin{table}[p!]
\centering
\begin{tabular}{rrrrr}
\toprule
\textbf{Entity Type} & \textbf{Train} & \textbf{Dev} & \textbf{Test} & \textbf{Test OOD} \\ \midrule
Numbers in addresses & 1,686          & 105          & 85            & 83                \\
Geographical names   & 773,901        & 41,759       & 38,903        & 33,001            \\
Institutions         & 601,129        & 28,380       & 33,119        & 52,938            \\
Media names          & 77,591         & 3,744        & 4,320         & 3,946             \\
Artifact names       & 159,122        & 7,550        & 7,174         & 25,204            \\
Personal names       & 747,686        & 36,783       & 37,712        & 65,950            \\
Time expressions     & 132,276        & 7,214        & 7,272         & 23,544            \\ \midrule
Total                & 2,493,391      & 125,535      & 128,585       & 204,666           \\ \bottomrule
\end{tabular}
\caption{Number of named entities in abstracts of SumeCzech's articles}
\label{tab:entities_abstract}
\end{table}

\subsection{Named Entity Annotations}
We train a model for named entity recognition in the Czech language to annotate SumeCzech by named entities. We selected the CoNLL-based extended \linebreak CNEC~2.0~\citep{11234/1-3493} as the training dataset, as it is the largest and most up-to-date Czech named entity recognition dataset. The advantage is that the dataset contains no nested entities, making the outputs easier to use for summarizers.

We selected SpaCy's NER model\footnote{\url{https://spacy.io/api/entityrecognizer}} \citep{spacy} because previous experiments by \citet{Muller2020Named} showed that SpaCy's NER model offers a good trade-off between performance, speed, and memory requirements. Speed and memory requirements might seem unimportant for our experiments because we can precompute the annotations. However, for the sake of practical usage, in which the labels have to be created as soon as possible once a new document for summarization arrives, we decided to take those properties into account too. The SpaCy's NER model achieved a 78.45 F-Score on the testing set of CoNLL-based extended CNEC 2.0. For comparsion, current state-of-the-art result on this dataset is 86.39 F-Score \citep{strakova-etal-2019-neural,vstvepan2020sumarizace}.

We applied the trained SpaCy's NER model to the text of SumeCzech's articles. The result was annotations in IOB2 format, one label for each word token. The NER found approximately 26M named entities in texts, 1M in headlines, and 3M in abstracts. (We do not use abstracts in our experiments. We present the numbers of named entities in the abstracts for completeness only.) We show the detailed statistics in \autoref{tab:entities_text}, \autoref{tab:entities_headline}, and \autoref{tab:entities_abstract}. We also counted the number of headlines without any named entity. We show the statistic in \autoref{tab:hedlines_without_entities}. We published the annotations\footnote{\url{http://hdl.handle.net/11234/1-3505}} to promote replication of results and to enable further research \citep{11234/1-3505}.

\begin{table}[h]
\centering
\begin{tabular}{lr}
\toprule
\textbf{Split} & \textbf{Percentage} \\ \midrule
Train          & 36.1\%              \\
Dev            & 35.4\%              \\
Test           & 35.7\%              \\
Test OOD       & 14.1\%              \\ \bottomrule
\end{tabular}
\caption{Percentage of headlines containing no named entity.}
\label{tab:hedlines_without_entities}
\end{table}

\section{Metrics}
We used the ROUGE\textsubscript{RAW} metric for evaluation. ROUGE\textsubscript{RAW} was proposed by \citet{straka2018sumeczech} as a language-agnostic variant of ROGUE \citep{lin2004rouge}. The original ROGUE metric automatically determines the quality of the generated summary by comparing it to a reference summary created by humans. There are two variants. ROUGE-N measures the overlap of N-grams between the generated and reference summaries. ROUGE-L looks at the longest common subsequence between the reference and the generated summaries. ROGUE calculates recall and is English-specific. It employs English stemmer, stop words, and synonyms.
 
ROUGE\textsubscript{RAW} does not need any stemmer, stop words, or synonyms, which makes it language independent. It measures recall, precision and F-score. It also has two variants ROUGE\textsubscript{RAW}-N and ROUGE\textsubscript{RAW}-L corresponding to the variants of the original ROGUE metric. We selected ROUGE\textsubscript{RAW}-1, ROUGE\textsubscript{RAW}-2, and ROUGE\textsubscript{RAW}-L to evaluate approaches that we propose.
 
\subsection{ROUGE\textsubscript{NE}}
 
Since we focus on the role of named entities in summarization, we propose a novel metric ROUGE\textsubscript{NE}. ROUGE\textsubscript{NE} measures the overlap of named entities between the reference and the generated summaries. This metric evaluates the ability of the model to transfer named entities to the summary.

Formally, let us denote the tokens of a true summary \(X\):
\[X = \{x_0, x_1, x_2, \ldots, x_n\},\]
where \(x_i\) are individual tokens of the summary. Let us denote the generated summary \(Y\) in a similar fashion:
\[Y = \{y_0, y_1, y_2, \ldots, y_m\}.\]
Next, we apply the named entity recognition algorithm on \(X\) and \(Y\). The result is entity annotations \(xe_i\) and \(ye_i\) for all tokens \(x_i\) and \(y_i\):
\[\{xe_0, xe_1, xe_2, \ldots, xe_n\},\]
\[\{ye_0, ye_1, ye_2, \ldots, ye_m\}.\]
The annotations can be divided into a set of annotations \(E_{NE}\), that mark entities, and annotations \(E_{¬NE}\), that do not mark entities. In the analogy of IOB format, the former are I and B annotations, and the latter are O annotations. For the calculation of ROUGE\textsubscript{NE} we select only tokens that are marked as entities. Formally, we select only the tokens of summaries \(X\) and \(Y\), for which its entity label is an element of \(E_{NE}\). The results are the \(X_e\) and \(Y_e\):
\[X_e = \{x_i\} \textrm{ for } i = 0\ldots n \textrm{ if } xe_i \in E_{NE},\]
\[Y_e = \{y_i\} \textrm{ for } i = 0\ldots m \textrm{ if } ye_i \in E_{NE}.\]

Next, we calculate the ROGUE precision and recall scores using the tokens of \(X_e\) and \(Y_e\) as follows:
\[precision = \frac{|X_e \cap Y_e|}{|X_e|},\]

\[recall = \frac{|X_e \cap Y_e|}{|Y_e|},\]
where \(|X_e|\) and \(|Y_e|\) denote the sizes of \(X_e\) and \(Y_e\). \(|X_e \cap Y_e|\) denotes the number of overlapping tokens between \(X_e\) and \(Y_e\). The resulting values are the precision and recall of ROUGE\textsubscript{NE}. We define the metrics to be equal to zero for summaries without any named entity.

\section{Methods}
The task we study is to create a one-sentence summary from the text of the article. The one-sentence summarization can be seen as the task to create a headline of the article. We use five baselines introduced by \citet{straka2018sumeczech}. Moreover, we propose one extractive method -- \emph{Named Entity Density} and two abstractive approaches, \emph{Seq2Seq} and \emph{Seq2Seq--NER}, for text summarization.

\subsection{Baselines}
We adopt the methods proposed by \citet{straka2018sumeczech} as a baseline. They propose four extractive and one abstractive methods for SumeCzech's task to create a headline out of the text of the article:
\begin{itemize}
    \item \emph{First}: unsupervised extractive method. It returns the first sentence of the article.
    \item \emph{Random}: unsupervised extractive method. It returns a random sentence from the article.
    \item \emph{TextRank}: unsupervised extractive method. It selects the most important sentence of the article using the TextRank \cite{mihalcea2004textrank} algorithm.
    \item \emph{clf-rf}: supervised extractive method. It selects the sentence that receives the highest score produced by the Random forest classifier. The classifier performs classification using vector representation of sentences. The vector representation consists of the sum of TF-IDF for each word normalized by the sentence length, length of the sentence, cohesion (distance from other sentences), the count of capitalized words in the sentence, the count of tokens that consist of digits, and the count of non-essential words that suggests the sentence relates to some other sentence.
    \item \emph{t2t}: supervised abstractive method. It uses a neural machine translation model proposed by \citet{vaswani2017attention} to generate a summary consisting of a novel sentence.
\end{itemize}

\subsection{Named Entity Density}
\emph{Named Entity Density} is our proposed unsupervised extractive method. It calculates the named entity density score for each sentence and selects the sentence with the highest score. The score is calculated in two steps. First, we apply a named entity recognition algorithm to all sentences. Next, we calculate the named entity density as a ratio of the number of tokens marked as a named entity to the total number of tokens in the sentence.

Formally, let us denote the article \(A\), for which we want to create a summary, as a set of sentences \(s_0 \ldots s_n\):
\[A = \{s_0, s_1, s_2, \ldots, s_n\}.\]
Each sentence \(s_i\) contains word tokens \(x_0 \ldots x_m\):
\[s_i = \{x_0, x_1, x_2, \ldots, x_m\}.\]
We apply the named entity recognition algorithm \(NER\) on each sentence \(s_i\) of the article \(A\):
\[NER(s_i) = \{e_0, e_1, e_2, \ldots, e_m\},\] that produces NER labels \(e_0 \ldots e_m\) for each token \(x_0 \ldots x_m\) of the sentence \(s_i\). We can divide the NER tokens into two sets: \(E_{NE}\) and \(E_{¬NE}\). Each NER token belongs into exactly one of those two sets. \(E_{NE}\) contains all NER tokens representing some entity type. \(E_{¬NE}\) contains all NER tokens that do not represent any entity type. In the analogy of IOB format, \(E_{NE}\) contains I and B tokens, and \(E_{¬NE}\) contains O tokens. Next, we calculate the named entity density \(NED\) for each sentence \(s_0 \ldots s_n\) of the article~\(A\). The \(NED\) is defined as:
\[NED(s_i)=\frac{|E_{NE}|}{|s_i|},\]
where \(|E_{NE}|\) denotes the number of tokens in the sentence \(s_i\) which NER algorithm marked as named entities and \(|s_i|\) is the number of all tokens \(x_0 \ldots x_m\) forming the sentence \(s_i\). We select the sentence \(s_i\) with the highest \(NED\) score as a summary of article~\(A\).
    
The intuition of the \emph{Named Entity Density} is that the sentence with the high \(NED\) score mentions the highest number of entities within the smallest text fragment. Such a sentence corresponds to the form of a report that is structured around concisely identifying and examining issues, events, or findings that have happened.
    
\subsection{Seq2Seq}
\label{section:Seq2Seq}
Seq2Seq is a supervised abstractive method that uses a Seq2Seq model with global attention. 
Formally, a Seq2Seq neural network models the conditional probability \(p(y|x)\) of translating a source text \(x=\{x_0,x_1,x_2, ..., x_n\}\) into a target text \linebreak \(y=\{y_0,y_1,y_2, ..., y_m\}\) \citep{luong2015effective}. The source text \(x\) is an article, \(y\) is a summary, and \(m < n\) in our case. The Seq2Seq neural network consists of an encoder and a decoder. The encoder creates a fixed-length vector representation \(r\) of the source text \(x\):
\[r = ENC(x).\] 
The encoder is usually a recurrent neural network with hidden states \(h^{ENC}_s\):
\[h^{ENC}_s = f^{ENC}(h^{ENC}_{s-1}, x_s),\] and the output of the encoder is its last hidden state:
\[ENC(x) = h^{ENC}_{n}\]
The \(f^{ENC}\) can be a vanila RNN \citep{rumelhart1985learning}, LSTM \citep{hochreiter1997long}, or GRU \citep{cho2014learning} unit. 

The decoder takes \(r\) as an input and generates a target text, one token at a time:
\[\log p(y|x) = \sum_{j=1}^{m}\log p(y_j|y_{<j}, r).\]
We can also represent the probability of generating a target word \(y_j\) as:
\[p(y_j|y_{<j}, r) = softmax(g(h^{DEC}_j)),\]
where \(g\) is a transformation function that generates vocabulary sized vector. The \(h^{DEC}_j\) is the output of recurrent neural network unit:
\[h^{DEC}_j=f^{DEC}(h^{DEC}_{j-1},y_{j-1}).\]
Function \(f^{DEC}\) can be a vanilla RNN, LSTM, or GRU unit like in the case of the encoder.
    
\begin{figure}[h!]
\centering
\begin{picture}(240,140)
\put(0,0){\includegraphics[width=0.65\textwidth]{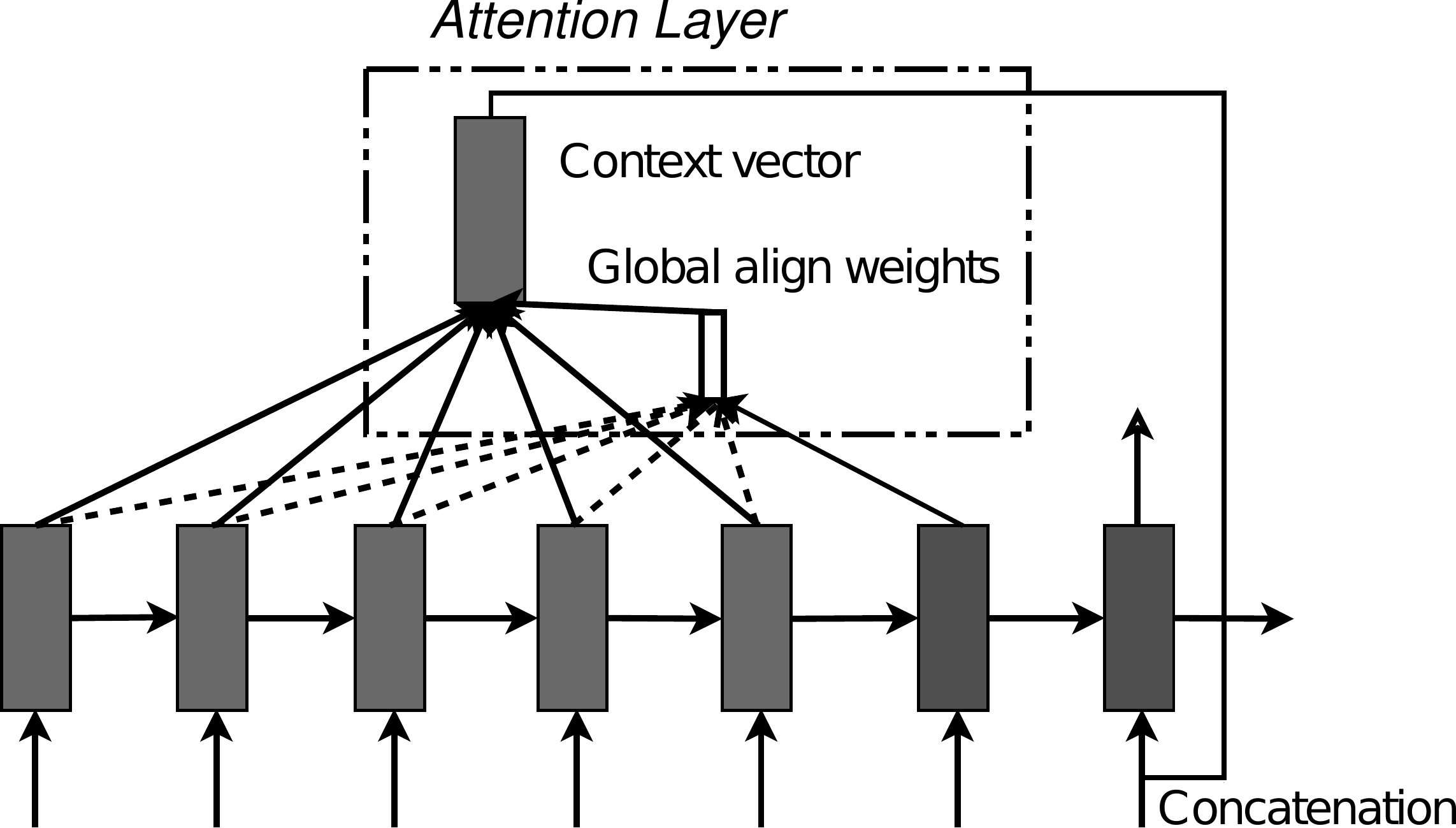}}
\put(169,8){$\tilde{y}_{j-1}$}
\put(185,74){$y_j$}
\put(65,100){$c_j$}
\put(123,75){$a_j$}
\put(155,56){$h^{DEC}_{j-1}$}
\put(0,56){$h^{ENC}_i$}
\end{picture}
	\caption{Seq2Seq with global attention mechanism. The figure is inspired by \citet{luong2015effective}.}
	\label{fig:attention}
\end{figure}
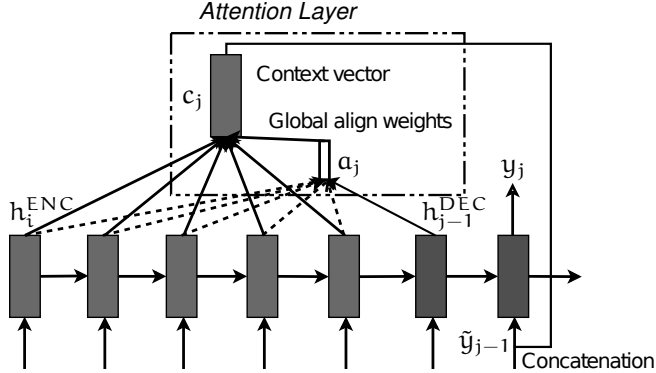

We add a global attention mechanism to the Seq2Seq neural network. The attention allows the network to focus on parts of the source text selectively during the target text generation. We illustrate the global attention mechanism in \autoref{fig:attention}.


The idea is to concatenate a source-side context vector \(c_j\) with the input vector of the encoder \(y_{j-1}\):
\[\tilde{y}_{j-1} = [c_j,y_{j-1}].\]
The vector \(\tilde{y}_{j-1}\) is fed into \(f^{DEC}\):
\[h^{DEC}_j=f^{DEC}(h^{DEC}_{j-1},\tilde{y}_{j-1}).\]

The context vector \(c_j\) is computed as a weighted average over all vectors of hidden states of the encoder \(h^{ENC}_s\):
\[c_j = \sum_{i=0}^{n}a_{j_i}\cdot h^{ENC}_i,\]
where \(a_{j_i}\) is the \(i^{th}\) element of the weight vector \(a_j\).
The \(a_j\) is calculated by the softmax function comparing each hidden state of the encoder \(h^{ENC}_s\) with the current hidden state of the decoder \(h^{ENC}_j\):
\[a_{j_i} = \frac{exp(score(h^{DEC}_{j-1},h^{ENC}_i))}{\sum_{s'}exp(score(h^{DEC}_{j-1},h^{ENC}_{s'}))}.\]
There are multiple definitions of the \(score\) function. We selected the \(general\) \(score\) function, defined as:
\[score(h_t,h_s) = h_t^\top W h_s.\]

\subsection{Seq2Seq--NER}
\emph{Seq2Seq--NER} is a supervised abstractive method that uses a Seq2Seq model with global attention and adds the NER feature encoded as one-hot encoded vector appended to input embedding vector.
Formally, the Seq2Seq neural network models the conditional probability \(p(y|x_{NER})\), where \(y\) is a target text \(y = \{y_0,y_1,y_2, ...,y_m\}\) and \(x_{NER}\) is a source sequence. The source sequence is a concatenation of a vector representation of the source token \(x_i\) and one-hot vector representation of entity type \(e_i\):
\[x_{NER} = \{[x_0,e_0],[x_1,e_1],[x_2,e_2], \ldots,[x_n,e_n]\}.\]
The rest of the model works in a similar fashion as a Seq2Seq model, that we described in the \autoref{section:Seq2Seq}.
To summarize, the difference between \emph{Seq2Seq} and \linebreak \emph{Seq2Seq--NER} models is that the latter has access to the named entity labels of the source words produced by the NER algorithm.

\section{Implementation Details}
We implemented the baseline methods \emph{first} and \emph{random} proposed by \citet{straka2018sumeczech}. We replicated results using ROUGE\textsubscript{RAW}-1, ROUGE\textsubscript{RAW}-2, and ROUGE\textsubscript{RAW}-L metrics, and additionally evaluated ROUGE\textsubscript{NE} metric.

For the proposed supervised methods, we used a modified implementation of a Seq2Seq model with global attention from the official PyTorch tutorial \citep{Seq2SeqPytorchTutorial}. The hidden sizes of the encoder and decoder were set to 256. The size of our vocabulary was 25,000. We used 300-dimensional fastText for embedding words. We used dropout 0.1 on the outputs of both RNNs. We trained our models until the validation loss started increasing, and we selected the weights having the lowest validation loss for evaluation.

We encountered problems with attention. The implementation in the tutorial used attention similar to \emph{concat global attention}. The tutorial suggested to copy hidden state of the decoder for each hidden state of the encoder, then concatenate the hidden states with the encoder outputs, and pass them to a linear layer to calculate energy. The reason was to utilize parallel nature of GPU. We used batch size 16. Therefore, the concatenated tensor would have 147,890,688 floating numbers in the case of the longest sentence. To train our model with attention without running out of memory on our GPU, we had to simplify the way attention was calculated. We used an approach similar to \emph{general global attention}. We used an affine transformation on the hidden state of the decoder to transform it into a 64-dimensional vector to calculate energy. We also affinely transformed the encoder's hidden states vectors of the same dimension by a linear layer. The transformed decoder hidden state was then used as a multiplier and broadcast over all encoder hidden states, making the calculation of energy much more memory efficient because the hidden state of the decoder did not have to be copied.

A limited vocabulary of the model led to many unknown words in the titles, and the model that used word tokenization learned to predict them. We had to forbid the model from predicting unknown tokens during evaluation to get meaningful titles.

For the \emph{Seq2Seq--NER} model, we encoded the entities using the IOB2 format. The format has one common \emph{outside} tag, and \emph{beginning} and \emph{inside} tags for each entity type. We encoded NER features into a one-hot vector for each word. The vector has 17 dimensions. Fourteen dimensions are reserved for the beginning and inside tags for each of the seven entity types our NER distinguishes. One dimension represents the outside of the entity tag. One represents padding, and one represents both the start and end of sequence symbols. We concatenated the NER feature vector with the embedding vector entering the encoder of the Seq2Seq.

\section{Results of Automatic Evaluation and Discussion}
We show the results of the evaluation in \autoref{tab:results}. First, we replicated the results of \emph{First} and \emph{Random} baselines reported in \citet{straka2018sumeczech}. Our results were on par with the reported Precision, Recall, and F-Score of Rogue\textsubscript{Raw}-1, Rogue\textsubscript{Raw}-2, \linebreak Rogue\textsubscript{Raw}-L. We additionally evaluated our proposed metric Rogue\textsubscript{NE} for comparison with other methods.

Next, we evaluated the proposed extractive method \emph{Named Entity Density} (NE~Density). Results of \emph{Named Entity Density} compared to the \emph{First}, a solid summarization baseline, especially in the news articles domain, are encouraging. \linebreak \emph{Named Entity Density} achieves only slightly worse results. Moreover, the achieved results are consistent between the test and the OOD test set. Additionally, as we will show in \autoref{section:examples}, \emph{Named Entity Density} produces summaries resembling the style of informationally concise reports.

We evaluated the \emph{Seq2Seq} and \emph{Seq2Seq--NER} on the Test set to compare those methods with the approaches proposed by \citet{straka2018sumeczech}. We can see that our proposed \emph{Seq2Seq} and \emph{Seq2Seq--NER} methods achieve better results on average by 80\% relatively in Precision and F-score compared to the best methods proposed by \citet{straka2018sumeczech}. Only \emph{Textrank} and \emph{First} achieve better results in Recall. The \emph{Seq2Seq--NER} achieved slightly better results than \emph{Seq2Seq}, which proves NER labels' usefulness for summarization. Although, it seems from the results of Rogue\textsubscript{NE} that the better score is not caused by the improved performance of using entities in the summaries.

We evaluated the \emph{Seq2Seq} and \emph{Seq2Seq--NER} on the OOD test set to learn how models cope with news articles from a domain never seen during training. The results are encouraging. Even though they show a drop in absolute values of metrics between Test and OOD test sets, the trend is the same. \emph{Seq2Seq} and \emph{Seq2Seq--NER} methods achieve the best results of all compared methods in Precision and F-score, and \emph{Seq2Seq--NER} has slightly better results than \emph{Seq2Seq}.

Finally, we take a look at the results in Rogue\textsubscript{NE}. We do not have results for \emph{Textrank} and \emph{Tensor2Tensor} because they were not reported in the work of \citet{straka2018sumeczech} and we did not implement the methods ourselves. However, it is clear from the rest of the results that even the recent state-of-the-art methods are struggling with the named entities in the summarization. 

\begin{table}[ht!]
\centering
\resizebox{\textwidth}{!}{%
\begin{tabular}{ll rrrrrrrrrrrr}
\toprule
\multirow{2}{*}{\textbf{Dataset}} & \multirow{2}{*}{\textbf{Method}} & \multicolumn{3}{c}{\textbf{Rogue\textsubscript{Raw}-1}}                                  & \multicolumn{3}{c}{\textbf{Rogue\textsubscript{Raw}-2}}                               & \multicolumn{3}{c}{\textbf{Rogue\textsubscript{Raw}-L}}                                  & \multicolumn{3}{c}{\textbf{Rogue\textsubscript{NE}}}                                 \\ \cmidrule(r){3-5}\cmidrule(lr){6-8}\cmidrule(lr){9-11}\cmidrule(l){12-14}
                                  &                                  & \textbf{P}             & \textbf{R}             & \textbf{F}             & \textbf{P}            & \textbf{R}            & \textbf{F}            & \textbf{P}             & \textbf{R}             & \textbf{F}             & \textbf{P}            & \textbf{R}            & \textbf{F}            \\ \midrule
\multirow{7}{*}{\textbf{Test}}    & First                            & 7.8                    & 14.6 & 9.4                    & 1.1                   & \textit{\textbf{2.3}}                   & 1.5                   & 6.7                    & 12.6                   & 8.1                    & 2.4                   & 2.7                   & 2.4                   \\
                                  & Random                           & 6.2                    & 11.0                   & 7.3                    & 0.5                   & 0.9                   & 0.6                   & 5.4                    & 9.5                    & 6.3                    & 1.8                   & 2.1                   & 1.8                   \\
                                  & Textrank                         & 6.0                    & \textit{\textbf{16.5}}                   & 8.3                    & 0.8                   & 0.6                   & 0.7                   & 5.0                    & \textit{\textbf{13.8}} & 6.9                    & -                     & -                     & -                     \\
                                  & Tensor2Tensor                    & 8.8                    & 7.0                    & 7.5                    & 0.8                   & 0.6                   & 0.7                   & 8.1                    & 6.5                    & 7.0                    & -                     & -                     & -                     \\ \cmidrule{2-14}
                                  & NE Density                    & 6.6                    & 10.7                   & 7.3                    & 0.8                   & 1.4                   & 0.9                   & 5.9                    & 9.4                    & 6.4                    & 1.5                   & 2.2                   & 1.6                   \\
                                  & Seq2Seq                          & 16.1                   & 14.1                   & 14.6                   & \textit{\textbf{2.5}}                   & 2.1                   & \textit{\textbf{2.2}}                   & 14.6                   & 12.8                   & 13.2                   & \textit{\textbf{5.3}}                   & \textit{\textbf{6.5}}                   & \textit{\textbf{5.6}}                   \\
                                  & Seq2Seq--NER                      & \textit{\textbf{16.2}} & 14.1                   & \textit{\textbf{14.7}} & \textit{\textbf{2.5}}                   & 2.1                   & \textit{\textbf{2.2}}                   & \textit{\textbf{14.7}} & 12.8                   & \textit{\textbf{13.3}} & 4.7                   & 6.0                   & 5.0                   \\ \midrule
\multirow{7}{*}{\textbf{OOD}}     & First                            & 7.0                    & 14.7 & 8.7                    & 1.4                   & \textit{\textbf{2.9}} & 1.7                   & 6.1                    & \textit{\textbf{12.8}} & 7.6                    & \textit{\textbf{1.4}} & 1.7          & \textit{\textbf{1.4}} \\
                                  & Random                           & 5.5                    & 10.9                   & 6.6                    & 0.7                   & 1.4                   & 0.8                   & 4.8                    & 9.5                    & 5.8                    & 0.9                   & 1.3                   & 1.0                   \\
                                  & Textrank                         & 5.8                    & \textit{\textbf{16.9}}                   & 8.1                    & 1.1                   & 3.4                   & 1.5                   & 5.0                    & 14.5                   & 6.9                    & -                     & -                     & -                     \\
                                  & Tensor2Tensor                    & 6.3                    & 5.1                    & 5.5                    & 0.5                   & 0.4                   & 0.4                   & 5.9                    & 4.8                    & 4.8                    & -                     & -                     & -                     \\ \cmidrule{2-14}
                                  & NE Density                      & 6.3                    & 11.4                   & 7.1                    & 1.3                   & 2.3                   & 1.4                   & 5.7                    & 10.2                   & 6.3                    & 1.0                   & \textit{\textbf{1.9}}                   & 1.1                   \\
                                  & Seq2Seq                          & 13.1                   & 11.8                   & 12.0                   & \textit{\textbf{2.0}}                   & 1.7                   & \textit{\textbf{1.8}}                   & 12.1                   & 11.0                   & 11.2                   & 1.0                   & 1.0                   & 1.0                   \\
                                  & Seq2Seq--NER                      & \textit{\textbf{13.7}} & 11.9                   & \textit{\textbf{12.4}} & \textit{\textbf{2.0}}                   & 1.7                   & \textit{\textbf{1.8}}                   & \textit{\textbf{12.6}} & 11.1                   & \textit{\textbf{11.4}} & 0.9                   & 0.9                   & 0.9                   \\ \bottomrule
\end{tabular}%
}
\caption{Results of automatic evaluation}
\label{tab:results}
\end{table}

\section{Examples}
\label{section:examples}

\begin{table}[p!]
\footnotesize
\begin{tabularx}{\textwidth}{l|X}
\hline
\textbf{Method}               & \textbf{Headline}                                                  \\ \hline
\multirow{2}{*}{\textbf{Gold}}         & Maloobchod v srpnu výrazně rostl          \\
                              & \textit{Retail trade grew significantly in August}   \\ \hline 
\multirow{2}{*}{\textbf{First}}        & Po očištění o sezónní a kalendářní vlivy rostl maloobchod meziročně o 4,2 procenta.                             \\
                              & \textit{After adjusting for seasonal and calendar effects, retail trade grew by 4.2 percent year on year.}     \\ \hline
\multirow{2}{*}{\textbf{NED}} & Podle Eurostatu vzrostly meziročně kalendářně očištěné maloobchodní tržby v celé Evropské unii o 2,2 procenta.                             \\
                              & \textit{According to Eurostat, calendar-adjusted retail sales rose by 2.2 percent year on year across the European Union.}     \\ \hline 
\hline
\multirow{2}{*}{\textbf{Gold}}         & Snoubenci zestárli, přibývá levnějších obřadů bez svatebčanů          \\
                              & \textit{The couple is getting old, there are more and more cheaper ceremonies without wedding guests}   \\ \hline 
\multirow{2}{*}{\textbf{First}}        & Stoupá počet sňatků bez svatebčanů, ve všední den, jen za přítomnosti svědků.                             \\
                              & \textit{The number of marriages without wedding guests is increasing, on weekdays, only in the presence of witnesses.}     \\ \hline
\multirow{2}{*}{\textbf{NED}} & Centrum metropole bude stále patřit k nejžádanějším místům pro oddávání, potvrdila Právu mluvčí Prahy 1 Veronika Blažková.                             \\
                              & \textit{The center of the metropolis will still be one of the most sought-after places for wedding, "Veronika Blažková, spokeswoman for Prague 1, confirmed to Právo.}     \\ \hline 
\hline
\multirow{2}{*}{\textbf{Gold}}         & Vranovskou přehradu znovu znečistila ropa, unikala ze sudů na dně          \\
                              & \textit{The Vranov dam was again polluted by oil, escaping from barrels at the bottom}   \\ \hline 
\multirow{2}{*}{\textbf{First}}        & Likvidace probíhá za odborné spolupráce pracovníků povodí Moravy a odboru životního prostředí.                             \\
                              & \textit{The liquidation takes place with the professional cooperation of the employees of the Moravia River Basin and the Department of the Environment.}     \\ \hline
\multirow{2}{*}{\textbf{NED}} & Starosta Vranova nad Dyjí se o ropě dozvěděl z tisku, což jej rozlítilo.                             \\
                              & \textit{The mayor of Vranov nad Dyjí learned about the oil from the press, which angered him.}     \\ \hline 
\hline
\multirow{2}{*}{\textbf{Gold}}         & Z Fondové bude Reaganova žena, doplní ji Oprah a Poslední skotský král          \\
                              & \textit{The Fond will be Reagan's wife, complemented by Oprah and the Last King of Scotland}   \\ \hline 
\multirow{2}{*}{\textbf{First}}        & Ve snímku s názvem The Butler (Majordomus) o správci v Bílém domě pracujícím pro několik amerických prezidentů by se v hlavní roli mohl podle časopsisu (the word \textit{časopsisu} is misspelled in the dataset) Variety objevit americký herec Forest Whitaker.                             \\
                              & \textit{According to Variety magazine, American actor Forest Whitaker could star in the film The Butler about a White House caretaker working for several US presidents.}     \\ \hline
\multirow{2}{*}{\textbf{NED}} & Amerického prezidenta Richarda Nixona si zřejmě zahraje John Cusak.                             \\
                              & \textit{US President Richard Nixon is likely to be played by John Cusack.}     \\ \hline 
\hline
\multirow{2}{*}{\textbf{Gold}}         & Zlatého ledňáčka na festivalu Finále Plzeň získal snímek Jako nikdy          \\
                              & \textit{Movie Jako nikdy won Golden Kingfisher at the Finale Pilsen festival}   \\ \hline 
\multirow{2}{*}{\textbf{First}}        & Letošní ročník festivalu Finále byl výjimečný tím, že poprvé soutěžily kromě českých také slovenské snímky.                             \\
                              & \textit{This year's Finale festival was exceptional in that, for the first time, in addition to Czech, Slovak films also competed}     \\ \hline
\multirow{2}{*}{\textbf{NED}} & Letošní ročník festivalu Finále Plzeň navštívilo od 27. dubna do 3. května 10 853 diváků.                             \\
                              & \textit{This year's Finale Plzeň festival was visited by 10,853 spectators from April 27 to May 3.}     \\ \hline 
\end{tabularx}
\caption{Examples of summarizations created by \emph{Named Entity Density}}
\label{tab:NERBaselineExamples}
\end{table}

We choose a few representative examples from the test and OOD test sets to show how different methods summarize. We also provide English translation for convenience. Only very simple automatic post-processing was done on the output of the proposed \emph{Seq2Seq} and \emph{Seq2Seq--NER} models. We filtered the start of the sentence and end of the sentence symbols, removed spaces before punctuation, stripped the text of any starting or ending space, and capitalized the first letter.

First, we present examples of summarization created by \emph{Named Entity Density} in \autoref{tab:NERBaselineExamples}. We do not divide the examples into test and OOD test sets because the generated summaries of both sets achieve comparative quality thanks to the fact that \linebreak \emph{Named Entity Density} is an unsupervised method. 

We can see that the selected sentences contain named entities. Those sentences are comprised of factual information. The sentences are always grammatically correct thanks to the fact that \emph{Named Entity Density} is an extractive approach. Even though the summaries created by \emph{First} can contain more entities in general, the summaries created by \emph{Named Entity Density} have a higher density of entities. We can see that the summaries created by \emph{Named Entity Density} revolve around \emph{to whom}, \emph{when}, \emph{where}, and \emph{what} happened. It closely resembles the style of reports that concisely identify and examine issues, events, or findings that have happened. 

Notice also that the sentences selected by \emph{Named Entity Density} are not the first sentences of the articles. We measured that the sentence selected as a summary by \linebreak \emph{Named Entity Density} differs from the sentence selected by \emph{First} in 93\% of SumeCzech's articles. Thus, we can use the summaries created by \emph{Named Entity Density} as an alternative version or reformulation of summaries created by method selecting the \emph{First} sentence of an article. This property is highly praised by voice applications like Alquist \citep{pichl2020alquist} or Emora \citep{finch2020emora}. Voice applications present news articles in a summary because users quickly lose focus as news articles are not intended to be read by synthetic voices. Initially, the voice application can present the first sentence of the article. Additionally, if the user requests to learn more, it can present the summary produced by \emph{Named Entity Density}.

We show the results of \emph{Seq2Seq} and \emph{Seq2Seq--NER} models for test and OOD test sets separately in \autoref{tab:testSetExamples} and \autoref{tab:OODSetExamples}. Both models generate novel sentences and incorporate entities into the generated summarizations successfully. We can see that despite the promising results of the automatic evaluation, a part of the outputs are not grammatically correct and contain repeated words.

\begin{table}[ht!]
\small
\begin{tabularx}{\textwidth}{l|X}
\hline
\textbf{Method}               & \textbf{Headline}                                                  \\ \hline
\multirow{2}{*}{\textbf{Gold}}         & Nejznámější Albánec může o stavbě mešity přemýšlet ve vězení          \\
                              & \textit{The most famous Albanian can think about building a mosque in prison}   \\ \hline 
\multirow{2}{*}{\textbf{Seq2Seq}}       & Soud potrestal za únos s lidmi                             \\
                              & \textit{The court punished for kidnapping with people}     \\ \hline
\multirow{2}{*}{\textbf{Seq2Seq--NER}} & Soud potvrdil tresty za pašování drog                             \\
                              & \textit{The court upheld the penalties for drug smuggling}     \\ \hline 
\hline
\multirow{2}{*}{\textbf{Gold}}         & Kriminalisté dopadli násilníka, který v lednu zneužil školáky z Orlové          \\
                              & \textit{Criminal investigators caught a rapist who abused schoolchildren from Orlová in January}   \\ \hline 
\multirow{2}{*}{\textbf{Seq2Seq}}        & Policie hledá muže, který v Ostravě znásilnil děti                             \\
                              & \textit{Police are looking for a man who raped children in Ostrava}     \\ \hline
\multirow{2}{*}{\textbf{Seq2Seq--NER}} & Policie hledá muže, který se v Ostravě, který se na něj                             \\
                              & \textit{Police are looking for a man in Ostrava who at him}     \\ \hline 
\hline
\multirow{2}{*}{\textbf{Gold}}         & Do Valtického Podzemí za divadlem místo vína          \\
                              & \textit{To the Valtice Underground for the theater instead of wine}   \\ \hline 
\multirow{2}{*}{\textbf{Seq2Seq}}        & Divadlo se v Brně otevře v Brně                             \\
                              & \textit{The theater in Brno will open in Brno}     \\ \hline
\multirow{2}{*}{\textbf{Seq2Seq--NER}} & V Brně otevřeli novou sezonu, divadlo se otevře návštěvníkům                             \\
                              & \textit{They have opened a new season in Brno, the theater will be open to visitors}     \\ \hline 
\end{tabularx}
\caption{Examples of summarizations from the Test set}
\label{tab:testSetExamples}
\end{table}

\begin{table}[ht!]
\small
\begin{tabularx}{\textwidth}{l|X}
\hline
\textbf{Method}               & \textbf{Headline}                                                  \\ \hline
\multirow{2}{*}{\textbf{Gold}}         & Havlova Asanace by sama asanaci potřebovala          \\
                              & \textit{Havel's Asanace itself would need sanitation}   \\ \hline 
\multirow{2}{*}{\textbf{Seq2Seq}}       & Havel se vrátil do divadla                             \\
                              & \textit{Havel returned to the theater}     \\ \hline
\multirow{2}{*}{\textbf{Seq2Seq--NER}} & Havel se s s Havlem Na zábradlí. Na hradě                             \\
                              & \textit{Havel with with Havel at Na zábradlí. On Castle}     \\ \hline 
\hline
\multirow{2}{*}{\textbf{Gold}}         & Hrad Bouzov nadchne cyklisty i zájemce o mučení a draky          \\
                              & \textit{Bouzov Castle will delight cyclists and those interested in torture and dragons}   \\ \hline 
\multirow{2}{*}{\textbf{Seq2Seq}}        & Hrady a zámky na hrad. Kde se můžete vidět i na hrad                             \\
                              & \textit{Castles and chateaux for the castle. Where you can yourself and castle}     \\ \hline
\multirow{2}{*}{\textbf{SeqSseq--NER}} & Na kole na hrad                             \\
                              & \textit{By bike to the castle}     \\ \hline 
\hline
\multirow{2}{*}{\textbf{Gold}}         & Filmy z Indie opět v Praze          \\
                              & \textit{Films from India again in Prague}   \\ \hline 
\multirow{2}{*}{\textbf{Seq2Seq}}        & V Indii se chystá na film o lásce                             \\
                              & \textit{There are preparations for a movie about love in India}     \\ \hline
\multirow{2}{*}{\textbf{Seq2Seq--NER}} & V Indii se vrací do Indie                             \\
                              & \textit{In India, he returns to India}     \\ \hline 
\end{tabularx}
\caption{Examples of summarizations from the OOD test set}
\label{tab:OODSetExamples}
\end{table}

\section{Conclusion}
This work explored the summarization of Czech news articles and influence of named entity labels for this task. We selected the SumeCzech dataset for our experiments. SumeCzech is over one million articles large dataset collected by \citet{straka2018sumeczech} from Czech news websites. We annotated SumeCzech by named entities by the SpaCy's NER. We published the annotations to promote replication of the results and to enable further research.

We used the methods introduced by \citet{straka2018sumeczech} as a baselines, namely \emph{First}, \emph{Random}, \emph{TextRank}, \emph{clf-rf}, and \emph{t2t}. We selected a task to create a headline out of the text of the article, which can be considered as a single sentence summary.

We proposed an extractive approach \emph{Named Entity Density} that selects a sentence with the highest ratio between the number of entities and length of the sentence as the summary of the article. The experiments showed that \emph{Named Entity Density} achieved nearly as good results as baseline selecting the first sentence of the article, which is a very hard baseline to beat, especially in the domain of news articles. Nevertheless, the summaries generated by \emph{Named Entity Density} demonstrated that the selected sentences reflect the style of reports concisely identifying \emph{to whom}, \emph{when}, \emph{where}, and \emph{what} happened. We proposed using a combination of \emph{Named Entity Density} and \emph{First}  summaries in voice applications. The voice application can initially present the first sentence of the article, and continue by follow-up created by \emph{Named Entity Density} if a user requests additional information.

Next, we proposed two abstractive approaches based on the Seq2Seq architecture. The first approach, \emph{Seq2Seq}, generates novel summaries using only tokens of the article's text. The second approach, \emph{Seq2Seq--NER}, additionally uses the named entity labels of each word as its input. Experiments showed that both proposed methods achieve better results than the methods proposed previously by \citet{straka2018sumeczech}. \emph{Seq2Seq--NER} improved the results over \emph{Seq2Seq} in automatic evaluation. This result demonstrated the usefulness of named entity labels for summarization. Furthermore, the results of the methods showed similar trends even on the out-of-domain test set.

Finally, we proposed a new metric, ROGUE\textsubscript{NE}, that measures the overlap of named entities between the true and generated summaries. The results show that the current state-of-the-art methods struggle with named entities in summarization, and there is a significant opportunity for further research.

\section*{Acknowledgements}
This research was supported by the Grant Agency of the Czech Technical University in Prague, grant No. SGS19/091/OHK3/1T/37.

\bibliography{mybib}


\correspondingaddress
\end{document}